\documentclass[11pt]{article}
\usepackage{times}

\usepackage{amsmath,graphicx}
\usepackage{amsmath,amssymb,amsfonts}
\usepackage[utf8x]{inputenc}
\usepackage{textgreek}
\usepackage{subfigure}
\usepackage{caption}
\usepackage{xcolor}
\usepackage{algorithmic}
\usepackage{graphicx}
\usepackage{textcomp}
\usepackage{xcolor}
\usepackage[numbers,sort]{natbib}

\usepackage{hyperref}

\usepackage{geometry}
\date{}
\begin{document}
\title{Learning of Image Dehazing Models for Segmentation Tasks}

\author{Sébastien de Blois, Ihsen Hedhli, Christian Gagné\\
Computer Vision and Systems Laboratory, Université Laval, Québec, QC Canada\\
\href{mailto:sebastien.de-blois.1@ulaval.ca}{sebastien.de-blois.1@ulaval.ca},
\href{mailto:ihsen.hedhli.1@ulaval.ca}{ihsen.hedhli.1@ulaval.ca}, 
\href{mailto:christian.gagne@gel.ulaval.ca}{christian.gagne@gel.ulaval.ca}}

\maketitle
%
%
%
\maketitle
\begin{abstract}
To evaluate their performance, existing dehazing approaches generally rely on distance measures between the generated image and its corresponding ground truth. Despite its ability to produce visually good images, using pixel-based or even perceptual metrics does not guarantee, in general, that the produced image is fit for being used as input for low-level computer vision tasks such as segmentation. To overcome this weakness, we are proposing a novel end-to-end approach for image dehazing, fit for being used as input to an image segmentation procedure, while maintaining the visual quality of the generated images. Inspired by the success of Generative Adversarial Networks (GAN), we propose to optimize the generator by introducing a discriminator network and a loss function that evaluates segmentation quality of dehazed images. In addition, we make use of a supplementary loss function that verifies that the visual and the perceptual quality of the generated image are preserved in hazy conditions. Results obtained using the proposed technique are appealing, with a favorable comparison to state-of-the-art approaches when considering the performance of segmentation algorithms on the hazy images.
\end{abstract}

Dehazing, Image segmentation, Deep neural network, Generative models 

\section{Introduction}
\label{sec:intro}
Images used for segmentation often suffer from poor weather conditions, including haze, snow, and rain. Nowadays, Deep Learning (DL) techniques are widely used to perform segmentation tasks. They require a large amount of training data covering different situations, including noisy images, in order to properly generalize the segmentation task. However, currently available datasets do not guarantee sufficient representativeness or even the presence of some meteorological conditions (e.g., haze) in their training data. Thus, testing the model with hazy images can hinder the performance for low-level computer vision algorithms such as segmentation~\cite{tan2008visibility}, even if this algorithm is known to be powerful in various circumstances. Therefore, restoring images from degraded observation under hazy conditions is a useful preprocessing step toward improving better segmentation. Several dehazing techniques have been extensively studied in the literature, most of them being based on physical models of image degradation, the problem being simplified to estimating the transmission map \cite{fattal2008single}. Such an estimation is made either by using a model (e.g., dark channel prior \cite{he2011single}) or through some learning approach (e.g., dehazenet \cite{cai2016dehazenet}). Other methods do not assume a physical model and try to build an end-to-end system for haze removal based on generative models (i.e., GAN) and restore original images directly from the hazy ones (e.g., \cite{li2018single, yang2018towards}).

Usually, the performance of dehazing is evaluated only with respect to some empirical measures. For instance, we note Structural Similarity Index (SSIM) and Peak signal to noise ratio (PSNR) \cite{ancuti2018ntire} as the most widely used measures found in the literature to evaluate the effectiveness of denoising methods. SSIM measures the similarity between two images, taking into account the similarity of the edges. PSNR is an indicator of the quality of the transmission of information. However, despite the performance of DL algorithms in dehazing in terms of SSIM and PSNR, there is no guarantee that they will produce images fit for being used as input to segmentation methods \cite{li2019benchmarking}.

This issue has been tackled by only a few papers found in the literature, which aim at reducing haze in order to improve segmentation quality. In \citet{li2017aod}, a dehazing algorithm for single hazy images was performed then fine-tuned for detection using a Fast-RCNN, with improvement reported on detection accuracy on the generated images. \citet{sakaridis2018semantic} proposed an end-to-end system for segmentation of a foggy scene, with training only on foggy dataset. Results show that the model was able to perform  segmentation well on hazy images. The proposed model differs from \citet{li2017aod} by adding a segmentation loss to the training procedure and not only a fine-tuning on segmentation and differ from \citet{sakaridis2018semantic} by being a preprocessing step toward segmentation of clean images and not an end-to-end segmentation technique for hazy images. More recently, \citet{ijcai2018} proposed a deep neural network solution that matches image denoising to computer vision tasks, and using the joint loss for updating only the denoising network via back-propagation. However, it assumes that only an independent and identically distributed Gaussian noise with zero mean is added to the original image as the noisy input image during training, which is not reflecting the usual noise we found in real-world problems. In the current paper, we are rather aiming at learning how to handle image degradation related to natural artifacts (i.e., haze).

The main contribution of the paper is a new DL-based dehazing system designed to take into account segmentation performance during training, for improved performance. We continue by detailing the methodology proposed (Sec.~\ref{sec:Methodology}), before reporting experiments and results (Sec.~\ref{sec:print}) validating the capacity of the proposed technique for an effective dehazing of images, usable for image segmentation. 

\section{Methodology}
\label{sec:Methodology}

\subsection{Single Image Dehazing}
\label{ssec:subhead}

Currently, several dehazing algorithms attempt to generate clean images using a generative model such as GAN \cite{goodfellow2014generative}, in which a generator attempted to fool a discriminator, providing most realistic fake sample possible. Specifically, \citet{li2018single} used a conditional GAN (CGAN) to perform single image dehazing. We are proposing a model derived from the Pix2Pix architecture \cite{isola2017image} for dehazing images, with the generator part of the CGAN composed of a downsampling, a residual, and an upsampling block structure inspired from \citet{johnson2016perceptual}. The loss function of the generator for single image dehazing is:
\begin{equation}\label{eq:h1}
    L_{generator}=L_{GAN}+\lambda_{1}\,L_{pixel}+\lambda_{2}\,L_{percep},
\end{equation}
where  $L_{GAN}$, $L_{pixel}$, and $L_{percep}$ are themselves loss functions on specific elements of the task and $\lambda_{i}$ are  weighting relative influence in a linear combination. $L_{GAN}$ is the loss function from \citet{isola2017image} used to generate fake images. $L_{pixel}$ is the reconstruction loss between the ground truth for dehazing (a.k.a. the real image) and the fake dehazed image, based on their individual pixel values, allowing the network to produce crisper images. $L_{percep}$ is the perceptual loss used for preserving important semantic elements of the image in the output of the generator. Indeed, it was shown \cite{li2018single, yang2018towards, johnson2016perceptual} that the use of perceptual loss improves the quality of the output by relying on the high-level representation features of a neural network (here a frozen VGG-16~\cite{simonyan2014very} is used), which are compared between the real image and the fake dehazed one. It is worth mentioning here that the same GAN discriminator loss function proposed in the original formulation of Pix2Pix is kept.

\subsection{Dehazing for Segmentation (DFS)}

\begin{figure*}[t] 
    \centering
    \subfigure{\includegraphics[width=0.90\textwidth]{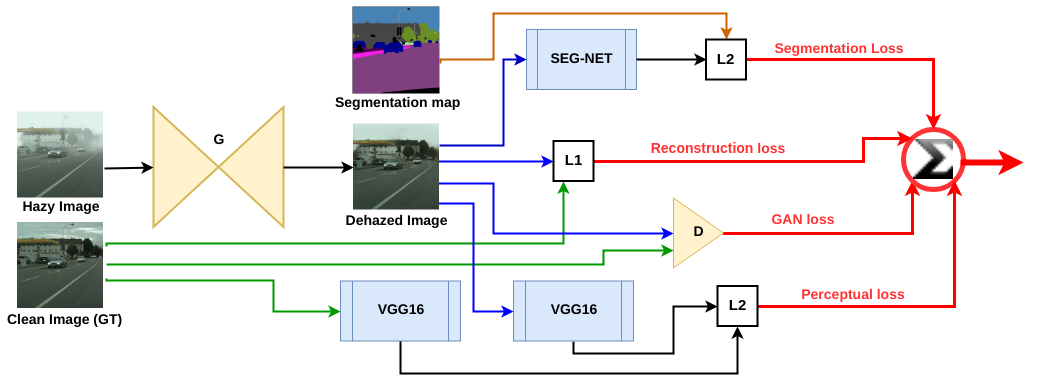}}
    \caption{Schema representing the DFS model}
    \label{fig:architecture}
\end{figure*}

The quality of segmentation relies on the input quality, which does not depend only on the acquisition devices but also on the environmental conditions such as haze or other weather conditions. Thus, removing these artifacts from the input images (e.g., by dehazing) can have an impact on segmentation quality. A dehazing method for segmentation purposes (referred to as DFS in this paper) is performed by using the following loss function to train the CGAN generator:
\begin{equation}\label{eq:h2}
    L_{generator}=L_{GAN}+\lambda_{1}\,L_{pixel}+\lambda_{2}\,L_{percep}+\lambda_{3}\,L_{seg}.
\end{equation}
Compared to Eq.~\ref{eq:h1}, it adds a new loss component, $L_{seg}$, which evaluates the impact of dehazing on the segmentation performance.

The architecture of the DFS model used is shown in Fig.~\ref{fig:architecture}. The generator network receives an image with haze as an input and gives a candidate of a dehazed image as the output. Similar to the single image dehazing model, the generator loss (Eq.~\ref{eq:h2}) is computed through $L_{GAN}$, $L_{pixel}$, and $L_{percep}$. The segmentation loss $L_{seg}$, is computed by placing the output of the generator (i.e., the dehazed image) into the segmentation network. The obtained segmentation map is then compared to the ground truth segmentation map, using the $L_2$ loss. Basically, the model tries at the same time to remove haze as much as possible while preserving, or even improving segmentation performance.

\begin{figure*}[t] 
    \small
  \centering
  \renewcommand*{\thesubfigure}{(\arabic{subfigure})}
    \subfigure[Hazy]{\includegraphics[scale=0.21]{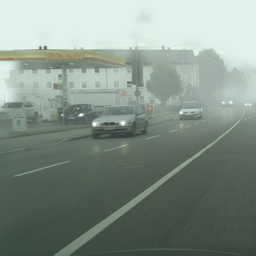}}\quad
    \subfigure[Dehaze]{\includegraphics[scale=0.21]{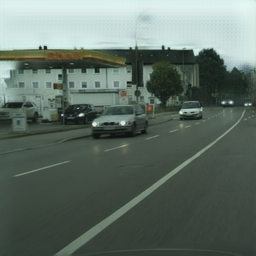}}\quad
     \subfigure[DFS]{\includegraphics[scale=0.21]{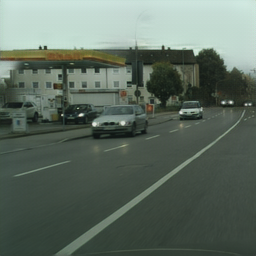}}\quad
    \subfigure[GT]{\includegraphics[scale=0.21]{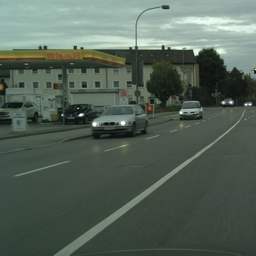}}\quad \vrule width 1pt \quad
    \subfigure[Hazy]{\includegraphics[scale=0.21]{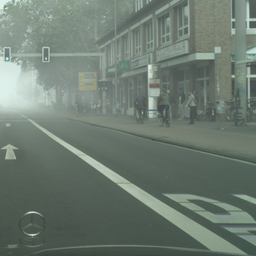}}\quad
    \subfigure[Dehaze]{\includegraphics[scale=0.21]{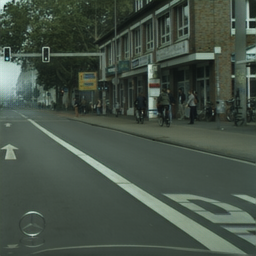}}\quad
     \subfigure[DFS]{\includegraphics[scale=0.21]{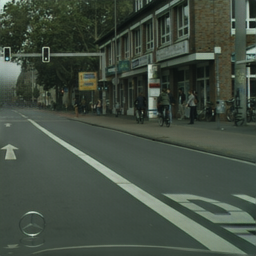}}\quad
    \subfigure[GT]{\includegraphics[scale=0.21]{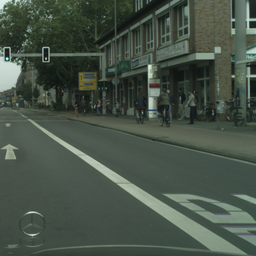}}\quad
    
    \subfigure[Hazy]{\includegraphics[scale=0.21]{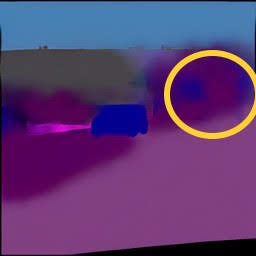}}\quad
    \subfigure[Dehaze]{\includegraphics[scale=0.21]{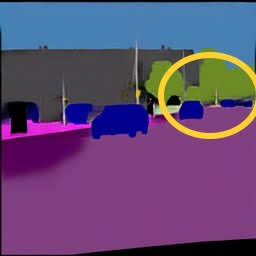}}\quad
     \subfigure[DFS]{\includegraphics[scale=0.21]{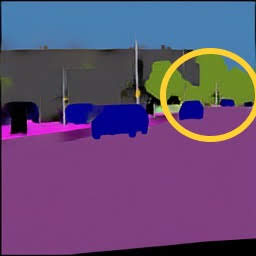}}\quad
    \subfigure[GT]{\includegraphics[scale=0.21]{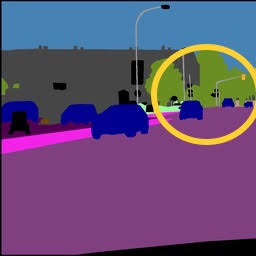}}\quad  \vrule width 1pt \quad 
    \subfigure[Hazy]{\includegraphics[scale=0.21]{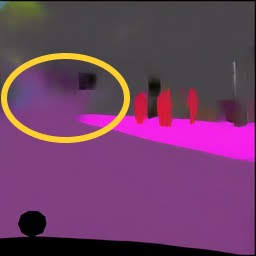}}\quad
    \subfigure[Dehaze]{\includegraphics[scale=0.21]{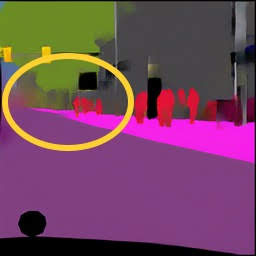}}\quad
     \subfigure[DFS]{\includegraphics[scale=0.21]{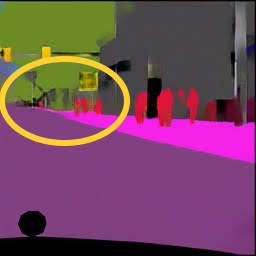}}\quad
    \subfigure[GT]{\includegraphics[scale=0.21]{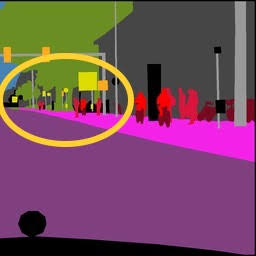}}\quad
    \hrule height 1pt
        \subfigure[Hazy]{\includegraphics[scale=0.21]{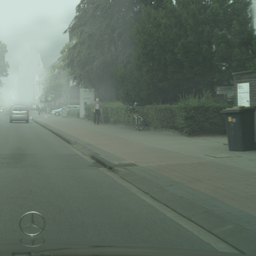}}\quad
    \subfigure[Dehaze]{\includegraphics[scale=0.21]{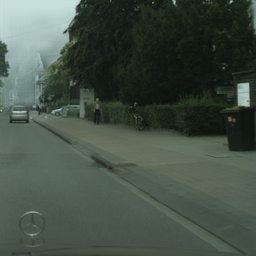}}\quad
     \subfigure[DFS]{\includegraphics[scale=0.21]{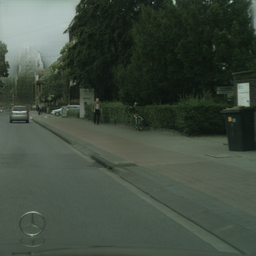}}\quad
    \subfigure[GT]{\includegraphics[scale=0.21]{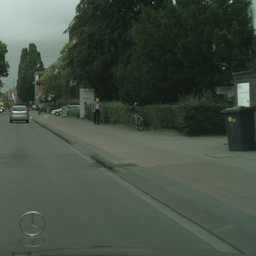}}\quad  \vrule width 1pt \quad
        \subfigure[Hazy]{\includegraphics[scale=0.21]{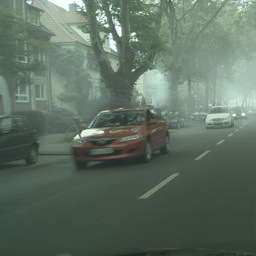}}\quad
    \subfigure[Dehaze]{\includegraphics[scale=0.21]{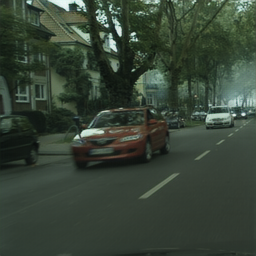}}\quad
     \subfigure[DFS]{\includegraphics[scale=0.21]{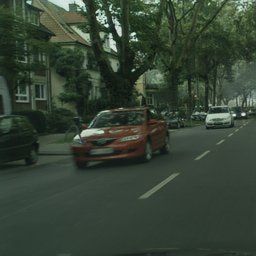}}\quad
    \subfigure[GT]{\includegraphics[scale=0.21]{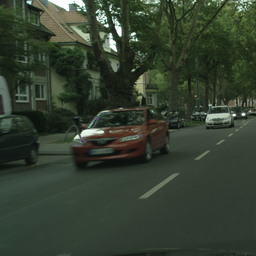}}\quad
    
    \subfigure[Hazy]{\includegraphics[scale=0.21]{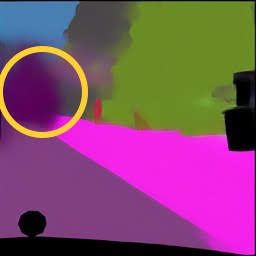}}\quad
    \subfigure[Dehaze]{\includegraphics[scale=0.21]{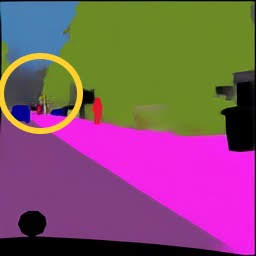}}\quad
     \subfigure[DFS]{\includegraphics[scale=0.21]{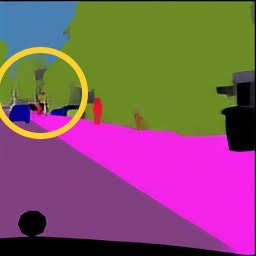}}\quad
    \subfigure[GT]{\includegraphics[scale=0.21]{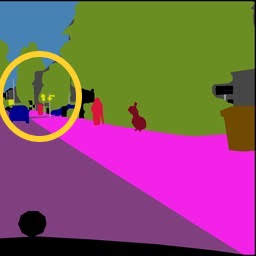}}\quad  \vrule width 1pt \quad
    \subfigure[Hazy]{\includegraphics[scale=0.21]{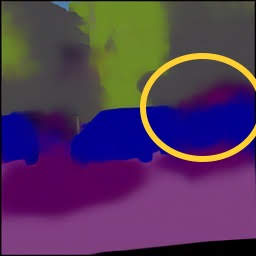}}\quad
    \subfigure[Dehaze]{\includegraphics[scale=0.21]{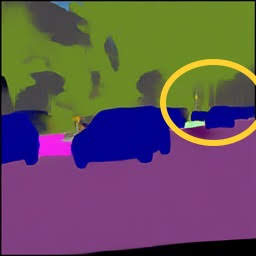}}\quad
     \subfigure[DFS]{\includegraphics[scale=0.21]{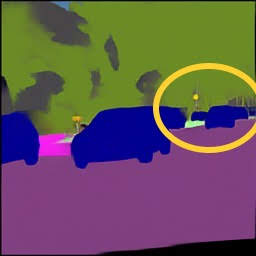}}\quad
    \subfigure[GT]{\includegraphics[scale=0.21]{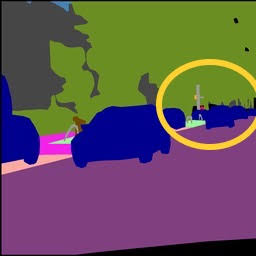}}\quad
    
  \caption{Four examples showing the performance of the proposed technique: Images 9-12, 25-28 and 29-32 present the problem of far away segmentation and images 13 to 16 show difficulties of far away human segmentation, they are the resulting segmentation map of respectively the images 1-4 for 9-13, 5-8 for 12-16, 17-20 for 25-28 and 21-24 for 29-32, segmentation is done using the SEG-NET, except for the GT segmentation map which is the ground truth, not the output of the SEG-NET.}
  \label{fig:images}
\end{figure*}

\section{Experiments and Results}
\label{sec:print}

\subsection{Single Image Dehazing}

\subsubsection{Dehazing Dataset}

The D-Hazy dataset is used for our experiments \cite{ancuti2016d}. D-Hazy contains 1449 pairs of synthetic hazy images  with ground truth based on the NYU Depth dataset \cite{silberman2012indoor}, which is composed of indoor images and their corresponding depth maps. The haze is computed from the depth map using physical models, providing realistic haze. The dataset is split 80\%/20\% between training and test partitions. Hyperparameters tuning is done through optimization over a random 80\%/20\% split of the training set. All the training set is used to infer the model with the selected hyperparameters. 

\subsubsection{Results for single image dehazing}

The proposed dehazing model was trained for 200 epochs using a batch size of 16 and Adam as the optimizer. Every image is resized to a resolution of 256x256. For the generator loss function (Eq.~\ref{eq:h1}), $\lambda_{1}$ and $\lambda_{2}$ are set as 10, following results obtained with a grid search over the validation set, which is made by testing 10 values of $\lambda$ between 1 and 50. Results obtained with the proposed technique achieve a PSNR of 17.89\,dB versus 15.55\,dB for the state-of-the-art \cite{yang2018towards} and a SSIM of 0.744 versus 0.77, a light degradation of 3.4\%. Table \ref{tab:ResultsNYU} presents the results of the proposed model versus multiple dehazing models, including the actual state-of-the-art, Disentangled Dehazing Network \cite{yang2018towards}. Because the test set used in our paper is different from the one used by \citet{yang2018towards}, we need to test algorithms appearing in the results of \cite{yang2018towards} with our test set. According to the results in Table 1, results obtained with our test set (marked with a *) show that it is of similar difficulty, maybe even harder, than the one used by \citet{yang2018towards}, with DCP and CycleGan performing less well on our test set. Our results also demonstrate empirically the capacity of the proposed model to remove haze from single images. 

\begin{table*}[t]
\centering
\caption{Results of the dehazing model on the NYU depth foggy dataset in comparison to the literature. DCP stands for Dark Channel Prior \cite{he2011single}, CAP for Color Attenuation Prior \cite{zhu2015fast}, NCP for Non-Local Color Prior \cite{berman2016non}, MSCNN for Multi-Scale CNN \cite{ren2016single}, DN for DehazeNet \cite{cai2016dehazenet}, CG for CycleGan \cite{zhu2017unpaired}, DDN for Disentangled Dehazing Network \cite{yang2018towards}, while Hazy is the original hazy image.  Results marked with a star (*) have been produced in our experiments, while the other non-marked results are reported from \citet{yang2018towards}.}
\label{tab:ResultsNYU}
\begin{tabular}{l|ccccccc|cccc}
\hline 
Metrics       & DCP \cite{he2011single} & CAP \cite{zhu2015fast} & NCP \cite{berman2016non} & MSCNN \cite{ren2016single} & DN \cite{cai2016dehazenet} & CG \cite{zhu2017unpaired} &  DDN \cite{yang2018towards} & Ours* & DCP*  & CG* & Hazy* \\ \hline
PSNR      & 10.98 & 12.78 & 13.05 & 12.27 & 12.84 & 13.39  & 15.55 & \textbf{\textit{17.89}} & \textit{9.12} & \textit{9.75} & \textit{3.67}   \\
SSIM            & 0.65 & 0.71 & 0.67 & 0.70& 0.72  & 0.52 & \textbf{0.77} & \textit{0.744} & \textit{0.52} & \textit{0.45} & \textit{0.20}   \\ \hline
\end{tabular}
\end{table*}

\subsection{DFS}

\subsubsection{Dataset for segmentation of hazy images}

The second set of experiments is conducted on the Cityscape dataset \cite{cordts2016cityscapes}, which includes a depth map, to simulate a realistic haze level while providing relatively complex scenes (i.e., a high traffic level in the images). For that purpose, we follow the methodology of \citet{sakaridis2018semantic} to generate realistic hazy image datasets. The final dataset is composed of $\approx$550 foggy images with the ground truth segmentation map in fine annotation. Around 50 images were randomly selected from the training set for the validation set while the 51 images proposed as a validation set for the foggy Cityscape \cite{sakaridis2018semantic} were used as a testing set.

\subsubsection{Training methodology and parameters}

In addition to its dehazing capability the proposed DFS model contains a segmentation loss which is computed using a segmentation model (referred to as SEG-NET). Except for the segmentation loss, this model is based on the same architecture as the model for single image dehazing. To train SEG-NET we use a subset of the Cityscape dataset not used to train the dehazing algorithm. The training was for 40 epochs using a batch size of 16 and Adam as the optimizer. The DFS model was trained for 100 epochs using a batch size of 8 and Adam as the optimizer. Here all of the images (size 2048x1024) are cropped into two squares (1024x1024) and resized to 256x256, cropping is necessary to keep the maximum amount of information and structure of the image before resizing it to a size and shape suited for this network. The parameters $\lambda_{1}$, $\lambda_{2}$ and $\lambda_{3}$ of Eq.\ref{eq:h2} have been set to 10, 10 and 5, respectively, by the application of a grid search using the same logic as for the single image dehazing, this time only on $\lambda_{3}$, both $\lambda_{1}$ and $\lambda_{2}$ are the same as for the single image dehazing. During the training of the DFS model, SEG-NET is frozen (i.e., not updated by gradient descent).

\subsubsection{Testing methodology}

During testing, segmentation results are reported on the same SEG-NET used for training the DFS model, and on another state-of-the-art segmentation model trained on Cityscape, DeepLabv3 \cite{chen2017rethinking}. The DeepLabv3 model was not used for training nor in validation, only for the final test reported here. Comparisons have been done between segmentation performances, after dehazing using models trained with and without segmentation loss, using foggy images and comparing with the ground truth non-hazy images. The DFS model and the single image dehazing model are trained using exactly the same training parameters.

Table \ref{tab:resultsSEGNET} presents the results of the PSNR, SSIM and MSE (mean squared error) with the hazy images, dehazed images, dehazed images with segmentation loss and the ground truth images, without haze, all using SEG-NET segmentation. Table \ref{tab:resultsDeepLabv3} presents the results with DeepLabv3, IoU is the intersection over union metric, ca is for categories, cl is for classes and iIoU refers to instance-level intersection-over-union metric. According to \citet{cordts2016cityscapes}, the IoU measure is biased toward object instances that cover a large image area. In street scenes, where there can be a strong scale variation, this can be problematic, especially for traffic participants. To address this problem it has been suggested to use the iIoU metric, where the contribution of each pixel is weighted by the ratio of the class average instance size to the size of the respective ground truth instance, only classes with instance annotations are included for this measure (classes: person, rider, car, truck, bus, train, motorcycle and bicycle, categories: human and vehicle).

\begin{table}
\centering
\caption{Results on the test set with SEG-NET segmentation (foggy Cityscape). For dehaze (single image dehazing) and dehaze with segmentation loss (DFS), results were done by averaging the best model obtained in validation on each on five different runs. Results in bold are the bests and results in italic are from the ground truth (GT).}
\label{tab:resultsSEGNET}
\begin{tabular}{c|c|c|c|c}
\hline 
Metrics       & Hazy & Dehaze & DFS & GT \\ \hline
PSNR            & 11.95 & 14.77 & \textbf{15.36} & \textit{15.45}  \\
SSIM            & 0.624 & 0.727 & \textbf{0.747} & \textit{0.748} \\ \hline
\end{tabular}
\end{table}
\begin{table}
\centering
\caption{Results on the test set with DeepLabv3 segmentation (foggy Cityscape). For dehaze (single image dehazing) and dehaze with segmentation loss (DFS), results were obtained by averaging the best model obtained in validation on each on five different runs. Results in bold are the bests and results in italic are from the ground truth (GT).}
\label{tab:resultsDeepLabv3}
\begin{tabular}{c|c|c|c|c}
\hline 
Metrics      & Hazy & Dehaze & DFS & GT \\ \hline
IoU-cl              & 0.556 & 0.553 & 0.557 & \textit{0.570}  \\
iIoU-cl             & 0.294  & 0.309 & \textbf{0.316} & \textit{0.340} \\
IoU-ca              & 0.745 & 0.774 & 0.775 & \textit{0.794} \\
iIoU-ca            & 0.518 & 0.554 & \textbf{0.569} & \textit{0.627} \\ \hline
\end{tabular}
\end{table}
 
\subsubsection{Discussion of results}

In the light of the results obtained with both SEG-NET and DeepLabv3 segmentation network, it appears that adding a loss for segmentation in the dehazing network significantly increases the accuracy of subsequent segmentation. 

With SEG-NET and DeepLabv3, results are always better when the output of DFS is used compared to dehazing models trained without segmentation loss. On average, the boost in segmentation with SEG-NET (PSNR/SSIM) from hazy to dehazed is around 20\%, while from dehazed to DFS, a 3.5\% gain is achieved. With DeepLabv3, the IoU metric differences are not significant for classes even with and without haze (0.556 versus 0.570), but there is a nice improvement for IoU with categories from hazy to dehaze (boost of 3.9\%), while a little gain (0.13\%), is observed from dehaze to DFS. Yet a significant improvement is observed with DeepLabv3 using segmentation loss with iIoU metric, the boost in segmentation for classes using iIoU from hazy to dehazed is of 5.1\%, while from dehazed to DFS the gain is 2.3\%. The boost in segmentation for categories using iIoU from hazy to dehazed is 7\% and for dehazed to DFS it reaches 2.7\%.

Looking at Fig.~\ref{fig:images}, subfigures 10, 11, 14, 15, 26, 27, 30, and 31, a significant improvement is perceptible for targets of interest located far away in the scene, especially with segmentation of cars and pedestrians, the largest differences being circled. The difference in segmentation performance between hazy (e.g., Fig.~\ref{fig:images}.9) and both dehazing techniques (e.g., Fig.~\ref{fig:images}.10-11) are big, adding the segmentation loss is giving a little boost making the results more similar to the ground truth (e.g., Fig.~\ref{fig:images}.12). Comparison  between normal dehazing and DFS (e.g., Fig.~\ref{fig:images}.22-23) shows more similar results, with DFS appearing to dehaze a little better for targets far away in the scene. The segmentation network also appears to be sensitive to elements that we usually don't notice when looking at the images.

\section{Conclusion}

In brief, this paper demonstrates the usefulness of including
segmentation loss in an end-to-end training of deep learning
models for dehazing. The learning-based dehazing model
is generated not just for denoising metrics, but also with an
optimization criterion aimed at achieving something useful for a
specific task, and with performance improvements that can be significant
in comparison to results obtained with an unguided approach. Moreover we can consider to boost even more the performance of DFS using directly an approximation of the IoU/iIoU measures for gradient descent \cite{berman2018lovasz}, which are better optimization measure than mean square error and similar. 

\section*{Acknowledgments}

This work was supported through funding from Mitacs, Thales Canada, and NSERC-Canada. We thank Annette Schwerdtfeger for proofreading this manuscript.

\bibliography{refs}
\bibliographystyle{unsrtnat}

\end{document}